\documentclass[twoside]{article}

\usepackage[accepted]{aistats2023}

\usepackage{amsmath,amsfonts,bm}

\def\eqref#1{equation~\ref{#1}}

\def\1{\bm{1}}

\def\vp{{\bm{p}}}

\def\vz{{\bm{z}}}

\DeclareMathAlphabet{\mathsfit}{\encodingdefault}{\sfdefault}{m}{sl}
\SetMathAlphabet{\mathsfit}{bold}{\encodingdefault}{\sfdefault}{bx}{n}

\newcommand{\E}{\mathbb{E}}

\newcommand{\KL}{D_{\mathrm{KL}}}

\newcommand{\rvar}[1]{\ensuremath{\mathit{#1}}\xspace}
\newcommand{\X}{\rvar{X}}
\newcommand{\Y}{\rvar{Y}}

\newcommand{\x}{\rvar{x}}
\newcommand{\y}{\rvar{y}}

\newcommand{\vv}[1]{\boldsymbol{#1}}

\newcommand{\xs}{\ensuremath{\vv{x}}\xspace}
\newcommand{\ys}{\ensuremath{\vv{y}}\xspace}
\newcommand{\zs}{\ensuremath{\vv{z}}\xspace}

\newcommand{\rvars}[1]{\ensuremath{\mathbf{#1}}\xspace}
\newcommand{\Xs}{\rvars{X}}
\newcommand{\Ys}{\rvars{Y}}

\newcommand{\prob}{P}
\newcommand\given[1][]{\:#1\vert\:}

\newcommand{\vars}{\ensuremath{\mathsf{vars}}}
\newcommand{\ch}{\ensuremath{\mathsf{in}}}

\usepackage{xspace}
\usepackage[dvipsnames]{xcolor}
\definecolor{sanae0}{rgb}{0.9312692223325372, 0.8201921796082118, 0.7971480974663592}
\definecolor{sanae1}{rgb}{0.8559578605899612, 0.6418993116910497, 0.6754191211563135}
\definecolor{sanae2}{rgb}{0.739734329496642, 0.4765280683170713, 0.5959617419736206}
\definecolor{sanae3}{rgb}{0.57916573903086, 0.33934576125314425, 0.5219003947563425}
\definecolor{sanae4}{rgb}{0.37894937987024996, 0.2224702044652721, 0.41140014301575434}
\definecolor{sanae5}{rgb}{0.1750865648952205, 0.11840023306916837, 0.24215989137836502}
\definecolor{ourspecialtextcolor}{rgb}{0.528, 0.471, 0.701}
\usepackage[colorlinks=true,linkcolor=Purple,citecolor=Blue]{hyperref}
\usepackage{url}       
\usepackage[capitalise,nameinlink,noabbrev]{cleveref}
\usepackage{graphicx} %

\usepackage{amssymb}

\usepackage{algorithm}
\usepackage{caption}
\usepackage{multicol}
\usepackage[noend]{algpseudocode}
\DeclareCaptionSubType*{algorithm}

\algrenewcommand{\algorithmiccomment}[1]{\bgroup\hfill//~#1\egroup}
\algrenewcommand{\Return}{\State\textbf{return}\ }
\algnewcommand{\Save}{\State\textbf{save}\ }
\algnewcommand{\Load}{\State\textbf{load}\ }

\usepackage{caption}
\usepackage{subcaption}

\usepackage{booktabs}

\usepackage{tikz}
\usetikzlibrary{circuits.logic.US}
\usetikzlibrary{shapes.misc}
\usetikzlibrary{shapes.arrows}
\usetikzlibrary{arrows,shapes.geometric}
\usetikzlibrary{external}
\usetikzlibrary{positioning}

\usepackage{pgfplots}
\pgfmathdeclarefunction{gaussian}{2}{%
  \pgfmathparse{1/(#2*sqrt(2*pi))*exp(-((x-#1)^2)/(2*#2^2))}%
}
\pgfmathdeclarefunction{gaussianmixture2}{5}{%
  \pgfmathparse{#5*(1/(#2*sqrt(2*pi))*exp(-((x-#1)^2)/(2*#2^2))) + (1-#5)*(1/(#4*sqrt(2*pi))*exp(-((x-#3)^2)/(2*#4^2)))}%
}

\newif\ifcomments
\ifcomments
    \newcommand{\kareem}[1]{\textbf{\textcolor{magenta}{{Kareem: #1}}}}
    \newcommand{\kw}[1]{\textbf{\textcolor{Peach}{{Kai-Wei: #1}}}}
    \newcommand{\guy}[1]{\textbf{\textcolor{blue}{{Guy: #1}}}}
\else
    \newcommand{\kareem}[1]{}
    \newcommand{\kw}[1]{}
    \newcommand{\guy}[1]{}
\fi

\usepackage[round]{natbib}

\DeclareMathOperator{\MI}{\mathtt{MI}}
\DeclareMathOperator{\SemStr}{\mathtt{SemanticStrengthening}}

\begin{document}

\twocolumn[

\aistatstitle{Semantic Strengthening of Neuro-Symbolic Learning}

\aistatsauthor{Kareem Ahmed\And Kai-Wei Chang \And  Guy Van den Broeck }

\aistatsaddress{ Computer Science Department\\UCLA\\ahmedk@cs.ucla.edu
\And  Computer Science Department\\UCLA\\kwchang@cs.ucla.edu
\And Computer Science Department\\UCLA\\guyvdb@cs.ucla.edu } ]

\begin{abstract}
Numerous neuro-symbolic approaches have recently been proposed typically with the goal of adding symbolic
knowledge to the output layer of a neural network.
Ideally, such losses maximize the probability that the neural network's predictions satisfy the 
underlying domain.
Unfortunately, this type of probabilistic inference is often computationally infeasible.
Neuro-symbolic approaches therefore commonly resort to fuzzy approximations of this probabilistic objective,
sacrificing sound probabilistic semantics, or to sampling which is very seldom feasible.
We approach the problem by first assuming the constraint decomposes conditioned on the features learned by the network.
We iteratively \emph{strengthen} our approximation, restoring the dependence between the constraints most responsible for degrading the quality of the approximation.
This corresponds to computing the mutual information between pairs of constraints conditioned on the network's learned features,
and may be construed as a measure of how well aligned the gradients of two distributions are.
We show how to compute this efficiently for tractable circuits.
We test our approach on three tasks: predicting a minimum-cost path in Warcraft, predicting a minimum-cost perfect matching, and solving Sudoku puzzles, observing that it improves upon the baselines while sidestepping intractability. %

\end{abstract}

\section{Introduction}
Neural networks have been established as excellent feature extractors, 
managing to learn intricate statistical features from large datasets.
However, without a notion of the \emph{symbolic rules} underlying any 
given problem domain, neural networks are often only able to achieve 
decent \emph{label-level} accuracy, with a complete disregard to the 
structure \emph{jointly} encoded by the individual labels.
These structures may encode, for example, a path in a graph, a matching of users to their preferences, or even the solution to a Sudoku puzzle.

Neuro-symbolic approaches \citep{RaedtIJCAI2020} hope to remedy 
the problem by injecting into the training process knowledge 
regarding the underlying problem domain, e.g. a Sudoku puzzle
is characterized by the uniqueness of the elements of every row,
column, and $3 \times 3$ square.
This is achieved by maximizing the probability allocated by the 
neural network to outputs satisfying the rules of the underlying 
domain.
Computing this quantity is, in general, a \#P-hard problem \citep{Valiant1979b}, which 
while tractable for a range of practical problems \citep{Xu18,Ahmed22nesyentropy},
precludes many problems of interest.

A common approach is to side step the hardness of computing the
probability \emph{exactly} by replacing logical operators with
their fuzzy t-norms, and logical implications with simple 
inequalities \citep{Grespan21, Krieken2020AnalyzingDF}.
This, however, does not preserve the sound probabilistic semantics
of the underlying logical statement: equivalent logic statements no
longer correspond to the same set of satisfying assignments, %
to different probability distributions, and consequently, vastly
different constraint probabilities.
On the other hand, obtaining a Monte Carlo estimate of the probability
\citep{Ahmed22pylon} is infeasible in exponentially-sized output spaces
where the valid outputs represent only a sliver of the distribution's 
support.

In this paper, starting from first principles, we derive a probabilistic approach to scaling probabilistic
inference for neuro-symbolic learning while retaining the sound semantics of the underlying logic.
Namely, we start by assuming that the probability of the constraint decomposes, conditioned on the network's learned features.
That is, we assume the events encoded by the logical formula to be \emph{mutually independent} given the learned features, and therefore, joint probability factorizes as a product of  probabilities.
This generalizes the prolific assumption that the probabilities of the variables are \emph{mutually-independent} conditioned on the network's learned features \citep{mullenbach2018explainable,Xu18,giunchiglia2020coherent} to events over arbitrary number of atoms.
This reduces the (often intractable) problem of probabilistically  satisfying the constraint, the validity of a Sudoku puzzle, to the (tractable) problem of probabilistically satisfying the individual local constraints, e.g. the uniqueness of the elements of a row, column, or square.
This, however, introduces inconsistencies: an assignment that satisfies one constraint might violate another, leading to misaligned gradients.
More precisely, for each pair of constraints, we are interested in the penalty incurred, in terms of modeling error, by assuming the constraints to be independent when they are in fact dependent, conditioned on the features learned by the neural network.
This corresponds exactly to the conditional mutual information, a quantity notoriously hard to calculate.
We give an algorithm for tractably computing the conditional mutual information, given that our constraints are represented as circuits satisfying certain structural properties.
Training then proceeds, where we interleave the process of learning the neural network, with the process of \emph{semantic strengthening}, where we iteratively tightening our approximation, using the neural network to guide us to which constraints need to be made dependent.

\definecolor{color1}{rgb}{0.50, 0.70, 1}
\definecolor{color2}{rgb}{0.10, 0.10, 0.1}

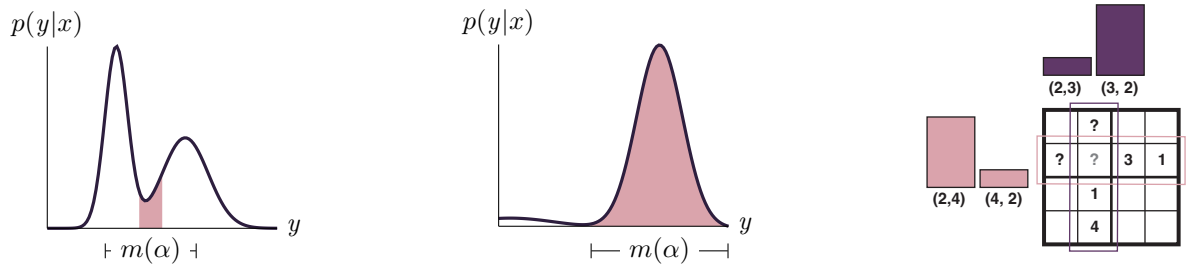
\begin{figure*}[t]
 \begin{subfigure}[b]{0.3\textwidth}
 \centering
 \begin{tikzpicture}
 \begin{axis}[
     no markers, domain=0:10, samples=100,
     axis lines*=left, xlabel=$\y$, ylabel=$p(\y|x)$,
     every axis y label/.style={at=(current axis.above origin),anchor=south},
     every axis x label/.style={at=(current axis.right of origin),anchor=west},
     height=4cm,
     xtick=\empty, ytick=\empty,
     enlargelimits=false, clip=false, axis on top,
     grid = major,
     ]
     \addplot [fill=sanae1, draw=none, domain=4.0:5.0] {gaussianmixture2(3,0.5,6,1,0.5)} \closedcycle;
     \addplot [very thick,sanae5] {gaussianmixture2(3,0.5,6,1,0.5)};
     \draw [yshift=-0.3cm, latex-latex, |-|](axis cs:2.5,0) -- node [fill=white] {$m(\alpha)$} (axis cs:6.5,0);
 \end{axis}
 \end{tikzpicture}
 \caption{Setting where satisfying assignments are only fraction of distribution support.}
 \end{subfigure}
 \hfill
\begin{subfigure}[b]{0.3\textwidth}
\centering
\begin{tikzpicture}
\begin{axis}[
    no markers, domain=0:10, samples=100,
    axis lines*=left, xlabel=$\y$, ylabel=$p(\y|x)$,
    every axis y label/.style={at=(current axis.above origin),anchor=south},
    every axis x label/.style={at=(current axis.right of origin),anchor=west},
    height=4cm,
    xtick=\empty, ytick=\empty,
    enlargelimits=false, clip=false, axis on top,
    grid = major,
    ]
    \addplot [fill=sanae1, draw=none, domain=4.0:10.0] {gaussianmixture2(0.5,3,7,1,0.1)} \closedcycle;
    \addplot [very thick,sanae5] {gaussianmixture2(0.5,2,7,1,0.1)};
    \draw [yshift=-0.3cm, latex-latex, |-|](axis cs:4.0,0) -- node [fill=white] {$m(\alpha)$} (axis cs:10,0);
\end{axis}
\end{tikzpicture}
\caption{A network allocating \emph{most} of probability mass to satisfying assignments.}
\end{subfigure}
\hfill
\begin{subfigure}[b]{0.3\textwidth}
\centering
\includegraphics[scale=0.5]{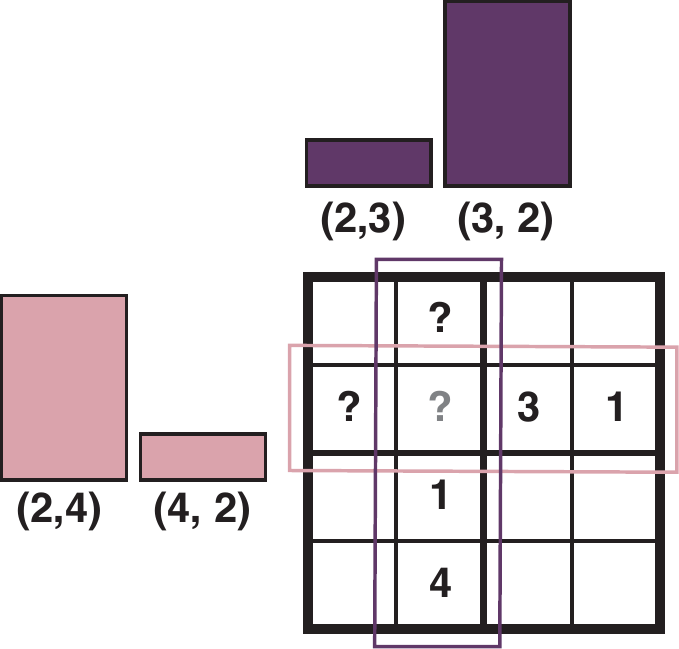}
\vspace{0.8em}
\caption{Distributions over empty entries of Sudoku row and col modeled separately.}
\end{subfigure}
\caption{%
Estimating the probability of a constraint using sampling can fail when, (a)
the set of satisfying assignments represents only a minuscule subset of the
distribution's support,
or, (b) when the network already largely satisfies the constraints, and consequently,
we are very unlikely to sample very low-probability assignments violating the constraint.
Using product t-norm, (c), to model the probability of satisfying constraints reduces the problem to satisfying the constraints locally, which can often lead to conflicting probabilities, and therefore, conflicting gradients. Here, e.g., according to the distribution over the Sudoku row, $3$ is the likely value of the cell in grey, where as, according to the distribution over the Sudoku column, $4$ is the likely value.
}
\label{fig:tnorm-failure}
\end{figure*}
We test our approach on three different tasks: predicting a minimum-cost path in a Warcraft terrain, predicting a minimum-cost perfect matching, as well as solving Sudoku puzzles, where we observe that our approach greatly improves upon the baselines all for a minuscule increase in computation time (our experiments are capped at ~2-3, and 7 seconds per iteration for Warcraft min-cost path, MNIST perfect matching, and Sudoku, respectively), thereby sidestepping the intractability of the problem. Our code is publiclt available at
\hyperlink{https://github.com/UCLA-StarAI/Semantic-Strengthening}{github.com/UCLA-StarAI/Semantic-Strengthening}.

\section{Problem Statement and Motivation}\label{sec:problem_statement}

We will start by introducing the notational choices used throughout the remainder of the paper, followed by a motivation of the problem.

We write uppercase letters ($\X$, $\Y$) for Boolean variables and lowercase letters ($\x$, $\y$) for their instantiation ($\Y=0$ or $\Y=1$).
Sets of variables are written in bold uppercase ($\Xs$, $\Ys$), and their joint instantiation in bold lowercase ($\xs$, $\ys$).
A literal is a variable ($\Y$) or its negation ($\neg \Y$).
A logical sentence ($\alpha$ or $\beta$) is constructed from variables and logical connectives ($\land$, $\lor$, etc.), and is also called a (logical) formula or constraint.
A state or world $\ys$ is an instantiation to all variables $\Ys$.
A state $\ys$ satisfies a sentence $\alpha$, denoted $\ys \models \alpha$, if the sentence evaluates to true in that world. A state $\ys$ that satisfies a sentence $\alpha$ is also said to be a model of $\alpha$.
We denote by $m(\alpha)$ the set of all models of $\alpha$.
The notation for states $\ys$ is used to refer to an assignment, the logical sentence enforcing the assignment, or the binary output vector capturing the assignment, as these are all equivalent notions.
A sentence $\alpha$ entails another sentence $\beta$, denoted $\alpha \models \beta$, if all worlds that satisfy $\alpha$ also satisfy $\beta$.

\paragraph{A Probability Distribution over Possible Structures}
Let $\alpha$ be a logical sentence defined over Boolean variables $\Ys = \{\Y_1,\dots,\Y_n\}$.
Let $\vp$ be a vector of probabilities for the same variables $\Ys$, where $\vp_i$ denotes the predicted probability of variable $\Y_i$ and corresponds to a single output of the neural network.
The neural network's outputs induce a probability distribution $\prob(\cdot)$ over possible states $\ys$ of $\Ys$:
\begin{equation}\label{eqn:pr_struct}
     \prob(\ys) = \prod_{i: \ys \models \Y_i} \vp_i \prod_{i: \ys \models \lnot \Y_i} (1 - \vp_i).
\end{equation}

\paragraph{Semantic Loss}\label{sec:semantic_loss}
 The semantic loss \citep{Xu18} is a function of the logical constraint $\alpha$ and a probability vector $\vp$. 
It quantifies how close the neural network comes to satisfying the constraint by computing the probability of the constraint under the distribution $\prob(\cdot)$ induced by $\vp$.
It does so by reducing the problem of probability computation to weighted model counting (WMC): summing up the models of $\alpha$, each weighted by its likelihood under $\prob(\cdot)$.
It, therefore, maximizes the probability mass allocated by the network to the models of $\alpha$
\begin{equation}
\label{eq:sloss}
\prob(\alpha)
= \mathbb{E}_{\ys \sim \prob}\left[ \1\{\ys \models \alpha\} \right] 
= \sum_{\ys \models \alpha} \prob(\ys). 
\end{equation}
Taking the negative logarithm recovers semantic loss.

Computing the above expectation is generally \#P-hard \citep{Valiant1979b}: there are potentially exponentially many models of $\alpha$. 
For instance, there are $6.67 \times 10^{21}$ valid $9 \times 9$ Sudokus \citep{Felgenhauer2005}, where as the number of valid matchings or paths in a $n \times n$ grid grows doubly-exponentially in the grid size \citep{STREHL2001}.

A common approach resorts to \emph{relaxing} the logical statements, replacing logical operators with their fuzzy t-norms, and implications with simple inequalities, and come in different flavors: Product~\citep{rocktaschel2015, li2019, asai2020}, G\"{o}del~\citep{minervini2017}, and \L{}ukasiewicz~\citep{Bach2017}, which differ only in their interpretation of the logical operators. \citet{Grespan21} offer a comprehensive theoretical, and empirical, treatment of the subject matter.

While attractive due to their tractability, t-norms suffer from a few major
drawbacks. First, they \emph{lose the precise meaning of the logical 
statement}, i.e. the satisfying and unsatisfying assignments of the relaxed
logical formula differ from those of the original logical formula. Second, the logic
is no longer consistent, i.e. logical statements that are otherwise equivalent
correspond to different truth values, as the relaxations are a function of 
their syntax rather than their semantics. Lastly, the relaxation sacrifices sound 
probabilistic semantics, unlike other approaches~\citep{Xu18, manhaeve2018}
where the output probability corresponds to the probability mass allocated to
truth assignments of the logical statement, the output probability 
has no sound probabilistic interpretation~\citep{Grespan21}.%

A slightly more benign relaxation~\citep{rocktaschel2015} only assumes that, 
for a constraint $\alpha = \beta_1 \land \ldots \land \beta_n$, a neural 
network $f(\cdot)$, and an input $\xs$, the events $\beta_i$ are mutually
independent conditioned on the features learned by the neural network.
That is, the probability of the constraint factorizes as $\prob(\alpha 
\given{f(\xs)}) = \prob(\beta_1 \given{f(\xs)}) \times \ldots \times 
\prob(\beta_n \given{f(\xs)})$.
This recovers the true probabilistic semantics of the logical statement 
when $\beta_1, \ldots, \beta_n$ are over disjoint sets of variables, 
i.e. $\forall_{i, j}$ $ \vars(\beta_i) \cap \vars(\beta_j) = \emptyset$ for $i \neq j$
and can otherwise
be thought of as a \emph{tractable} approximation, the basis of which
is the neural network's ability to sufficiently encode the dependencies
shared between the constraints, rendering them conditionally independent
given the learned features.
That is assuming the neural network makes almost-deterministic predictions
of the output variables given the embeddings.
However, even assuming the true function being learned is deterministic,
there is still the problem of an imperfect embedding giving probabilistic 
predictions whereby clauses are dependent.

The above relaxation reduces the \emph{intractable} problem of satisfying
the global constraint to the \emph{tractable} problem of satisfying the local
constraints, and can therefore often lead to \emph{misaligned gradients}.
Consider cell $(1,1)$ of the Sudoku in \cref{fig:tnorm-failure}.
Consider the two constraints asserting that the elements of row $2$ and that 
the elements of column $2$ are unique, and assume the probability distribution 
induced by the network over row and column assignments are as shown in 
\cref{fig:tnorm-failure}, right.
This leads to opposing gradients for cell $(1,1)$: On the one hand, the gradient 
from maximizing the probability of the column constraint pushes it to $2$, whereas
the gradient from maximizing the probability of the row constraint pushes it to $4$.
The problem here stems from modeling as independent two constraints that are strongly
coupled, so much so that the value of one determines the value of the other.

Recently, \citet{Ahmed22pylon} proposed using sampling to obtain a Monte Carlo
estimate of the probability of the constraint being satisfied.
This offers the convenience of specifying constraints as PyTorch functions,
as well as accommodating non-differentiable elements in the training pipeline
of the constraint, especially in cases where the training pipeline includes non-differentiable elements.
However, when problems are intractable, this is often accompanied by a state space that is combinatorial in size, meaning that the probability of sampling a valid structure drops precipitously as a function of the size of the state space, making it near impossible to obtain any learning signal, as almost all the sampled states will necessarily violate our constraint.
The same applies when the constraint is almost satisfied, meaning we never sample
low-probability assignment that violate the constraint.

That is not to mention the downfalls of gradient estimators: the gradient estimator employed by \citet{Ahmed22pylon} is the REINFORCE gradient estimator, which while unbiased in the limited of many samples, exhibits variances that makes it very hard to learn. Even gradient estimators that do not exhibit this problem of variance, trade off variance for bias, making it unlikely to obtain the true gradient.

\section{Semantic Strengthening}
We are interested in an approach that, much like the approaches
discussed in \cref{sec:problem_statement} is tractable, but retains sound 
probabilistic semantics, and yields a non-zero gradient when the constraint
is locally, or globally, violated.

Let our constraint $\alpha$ be given by a conjunctive normal form (CNF), 
$\alpha = \beta_1 \land \ldots \land \beta_n$.
We start by assuming that, for a neural 
network $f(\cdot)$, and an input $\xs$, the clauses $\beta_i$
are mutually independent conditioned on the features learned 
by the neural network i.e. the probability of the constraint 
factorizes as $\prob(\alpha \given{f(\xs)}) = \prob(\beta_1 
\given{f(\xs)}) \times \ldots \times \prob(\beta_n \given{
f(\xs)})$, where the probability of each of the clauses, 
$\prob(\beta_i)$, can be computed tractably.
This recovers the true probabilistic semantics of the logical statement 
when $\beta_1, \ldots, \beta_n$ are over disjoint sets of variables, 
i.e. $\forall_{i, j}$ $ \vars(\beta_i) \cap \vars(\beta_j) = \emptyset$ for $i \neq j$,  and can otherwise
be thought of as a \emph{tractable} approximation, the basis of which
is the neural network's ability to sufficiently encode the dependencies
shared between the constraints, rendering them conditionally independent
given the learned features, again, assuming the true function is deterministic,
with no inherent uncertainty.

The above approximation is semantically sound in the sense that, the probability
of each term $\prob(\beta_i)$ accounts for all the truth assignment of the clause
$\beta_i$.
It is also guaranteed to yield a semantic loss value of $0$,
and therefore a zero gradient
if and only if all the clauses, $\beta_i$, are satisfied.

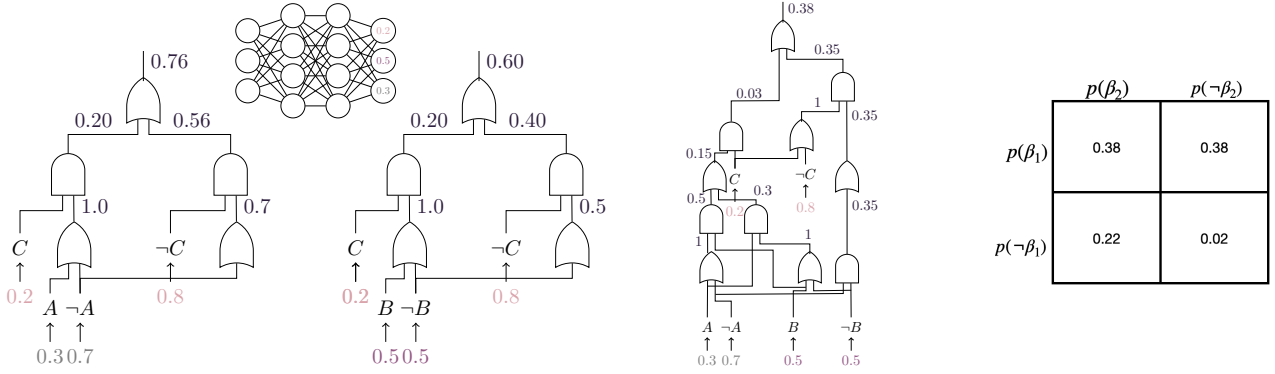
\begin{figure*}[t!]
\begin{subfigure}[b]{0.18\textwidth}
\centering
\scalebox{.8}{
\begin{tikzpicture}[circuit logic US]
\node (or1) [or gate, inputs=nn, rotate=90, scale=0.9] at (4.25,7) {};
\node (output) at (4.25, 8.0) {};
\draw (or1) -- (output) node[pos=0.2, above right, color=sanae5] {$0.76$};

\node (and1) [and gate, inputs=nn, rotate=90, scale=0.9] at (3,5.8) {};
\node (and2) [and gate, inputs=nn, rotate=90, scale=0.9] at (5.7,5.8) {};

\node (c) at (2.2,4.6) {$C$};
\node (cval) at (2.2,3.8) {$\color{sanae1}0.2$};
\draw[->] (cval) edge (c);
\node (nc) at (4.7,4.6) {$\neg C$};
\node (ncval) at (4.7,3.8) {$\color{sanae1}0.8$};
\draw[->] (ncval) edge (nc);
\node (or2) [or gate, inputs=nn, rotate=90, scale=0.9] at (3.1,4.6) {};
\node (or3) [or gate, inputs=nnn, rotate=90, scale=0.9] at (5.8,4.6) {};

\node (a1) at (2.7,3.6) {$A$};
\node (a1val) at (2.7,2.8) {$\color{gray}0.3$};
\draw[->] (a1val) edge (a1);

\node (na1) at (3.2,3.6) {$\neg A$};
\node (na1val) at (3.2,2.8) {$\color{gray}0.7$};
\draw[->] (na1val) edge (na1);

\draw (or1.input 1) -- ++ (down: 0.25) -| (and1) node[pos=-0.2, above right, color=sanae5] {$0.56$};
\draw (or1.input 2) -- ++ (down: 0.25) -| (and2) node[pos=-0.5, above right, color=sanae5] {$0.20$};

\draw (and1.input 1) -- ++ (down: 0.25) -| (c);
\draw (and1.input 2) -- (or2) node[pos=0.4, right, color=sanae5] {$1.0$};

\draw (and2.input 1) -- ++ (down: 0.25) -| (nc);
\draw (and2.input 2) -- (or3) node[pos=0.4,  right, color=sanae5] {$0.7$};

\draw (or2.input 1) -- ++ (down: 0.25)  -| (a1);
\draw (or2.input 2) edge (na1);

\draw (or3.input 2) -- ++ (down: 0.25)  -| (na1);
\end{tikzpicture}
}
\end{subfigure}
\begin{subfigure}[b]{0.07\textwidth}
\scalebox{.4}{
\begin{tikzpicture}[baseline=-250pt]
\node[circle,draw,minimum size=0.8cm] (a1) at (0,0.5) {};
\node[circle,draw,minimum size=0.8cm] (a2) at (0,1.5) {};
\node[circle,draw,minimum size=0.8cm] (a3) at (0,2.5) {};

\node[circle,draw,minimum size=0.8cm] (b1) at (1.5,0) {};
\node[circle,draw,minimum size=0.8cm] (b2) at (1.5,1) {};
\node[circle,draw,minimum size=0.8cm] (b3) at (1.5,2) {};
\node[circle,draw,minimum size=0.8cm] (b4) at (1.5,3) {};

\foreach \i in {1,...,3} {
    \foreach \j in {1,...,4} {
        \draw (a\i) -- (b\j);
    }
}

\node[circle,draw,minimum size=0.8cm] (c1) at (3.0,0) {};
\node[circle,draw,minimum size=0.8cm] (c2) at (3.0,1) {};
\node[circle,draw,minimum size=0.8cm] (c3) at (3.0,2) {};
\node[circle,draw,minimum size=0.8cm] (c4) at (3.0,3) {};

\foreach \i in {1,...,4} {
    \foreach \j in {1,...,4} {
        \draw (b\i) -- (c\j);
    }
}

\node[circle,draw,minimum size=0.8cm] (d1) at (4.5,0.5) {$\color{gray}0.3$};
\node[circle,draw,minimum size=0.8cm] (d2) at (4.5,1.5) {$\color{sanae3}0.5$};
\node[circle,draw,minimum size=0.8cm] (d3) at (4.5,2.5) {$\color{sanae1}0.2$};

\foreach \i in {1,...,4} {
    \foreach \j in {1,...,3} {
        \draw (c\i) -- (d\j);
    }
}
\end{tikzpicture}
}
\end{subfigure}
\begin{subfigure}[b]{0.16\textwidth}
\centering
\scalebox{.8}{
\begin{tikzpicture}[circuit logic US]
\node (or1) [or gate, inputs=nn, rotate=90, scale=0.9] at (4.25,7) {};
\node (output) at (4.25, 8.0) {};
\draw (or1) -- (output) node[pos=0.2, above right, color=sanae5] {$0.60$};

\node (and1) [and gate, inputs=nn, rotate=90, scale=0.9] at (3,5.8) {};
\node (and2) [and gate, inputs=nn, rotate=90, scale=0.9] at (5.7,5.8) {};

\node (cval) at (2.2,3.8) {$\color{sanae1}{0.2}$};
\draw[->] (cval) edge (c);

\node (c) at (2.2,4.6) {$C$};
\node (cval) at (2.2,3.8) {$\color{sanae1}0.2$};
\draw[->] (cval) edge (c);
\node (nc) at (4.7,4.6) {$\neg C$};
\node (ncval) at (4.7,3.8) {$\color{sanae1}0.8$};
\draw[->] (ncval) edge (nc);
\node (or2) [or gate, inputs=nn, rotate=90, scale=0.9] at (3.1,4.6) {};
\node (or3) [or gate, inputs=nnn, rotate=90, scale=0.9] at (5.8,4.6) {};

\node (a1) at (2.7,3.6) {$B$};
\node (a1val) at (2.7,2.8) {$\color{sanae3}0.5$};
\draw[->] (a1val) edge (a1);

\node (na1) at (3.2,3.6) {$\neg B$};
\node (na1val) at (3.2,2.8) {$\color{sanae3}0.5$};
\draw[->] (na1val) edge (na1);

\draw (or1.input 1) -- ++ (down: 0.25) -| (and1) node[pos=-0.2, above right, color=sanae5] {$0.40$};
\draw (or1.input 2) -- ++ (down: 0.25) -| (and2) node[pos=-0.5, above right, color=sanae5] {$0.20$};

\draw (and1.input 1) -- ++ (down: 0.25) -| (c);
\draw (and1.input 2) -- (or2) node[pos=0.4, right, color=sanae5] {$1.0$};

\draw (and2.input 1) -- ++ (down: 0.25) -| (nc);
\draw (and2.input 2) -- (or3) node[pos=0.4,  right, color=sanae5] {$0.5$};

\draw (or2.input 1) -- ++ (down: 0.25)  -| (a1);
\draw (or2.input 2) edge (na1);

\draw (or3.input 2) -- ++ (down: 0.25)  -| (na1);

\end{tikzpicture}
}
\end{subfigure}
\hspace{1em}
\begin{subfigure}[b]{0.32\textwidth}
\centering
\scalebox{.55}{
\begin{tikzpicture}[circuit logic US]
\node (output) at (4.25, 8.0) {};
\node (or1) [or gate, inputs=nn, rotate=90, scale=0.9] at (4.25,7) {};
\node (output) at (4.25, 8.0) {};
\draw (or1) -- (output) node[pos=0.2, above right, color=sanae5] {$0.38$};
\draw (or1.input 1) -- ++ (down: 0.25)  -- ++ (down: 1.0) -- ++ (down: 0.0) -- (3, 5.48) -- (3, 4.98) node[pos=0.1, above right, color=sanae5] {$0.03$};
\draw (or1.input 2) -- ++ (down: 0.25) -| (and2) node[pos=0.2, above right, color=sanae5] {$0.35$};

\node (and1) [and gate, inputs=nn, rotate=90, scale=0.9] at (3,4.6) {};
\node (or2) [or gate, inputs=nn, rotate=90, scale=0.9] at (2.6,3.6) {};
\node (c) at (3.09,3.6) {$C$};
\node (nc) at (4.8,3.7) {$\neg C$};

\node (cval) at (3.09,2.8) {$\color{sanae1}0.2$};
\draw[->] (cval) edge (c);

\node (ncval) at (4.8,2.9) {$\color{sanae1}0.8$};
\draw[->] (ncval) edge (nc);

\draw (and1.input 2) edge (c);
\draw (and1.input 1) -- ++ (down: 0.05) -| (or2) node[pos=0.4, left, color=sanae5] {$0.15$};

\node (and2) [and gate, inputs=nn, rotate=90, scale=0.9] at (5.7,5.8) {};
\node (or3) [or gate, inputs=nn, rotate=90, scale=0.9] at (4.7,4.6) {};
\draw (or3.input 1) -- + (down: 0.25) -| (c);
\draw (or3.input 2) edge (nc);

\node (or4) [or gate, inputs=nnn, rotate=90, scale=0.9] at (5.8,3.6) {};
\draw (and2.input 1) -- ++ (down: 0.25) -| (or3) node[pos=0.4, above right, color=sanae5] {$1$};
\draw (and2.input 2) -- (or4) node[pos=0.4, above right, color=sanae5] {$0.35$};

\node (and3) [and gate, inputs=nn, rotate=90, scale=0.9] at (5.8,1.4) {};
\draw (or4.input 2) -- (and3) node[pos=0.4, above right, color=sanae5] {$0.35$};

\node (b) at (4.5, 0) {$B$};
\node (nb) at (5.9,0) {$\neg B$};

\node (bval) at (4.5,-0.8) {$\color{sanae3}0.5$};
\draw[->] (bval) edge (b);

\node (nbval) at (5.9,-0.8) {$\color{sanae3}0.5$};
\draw[->] (nbval) edge (nb);

\node (a) at (2.42, 0) {$A$};
\node (na) at (3.0,0) {$\neg A$};

\node (aval) at (2.42,-0.8) {$\color{gray}0.3$};
\draw[->] (aval) edge (a);

\node (naval) at (3.0,-0.8) {$\color{gray}0.7$};
\draw[->] (naval) edge (na);

\draw (and3.input 2) edge (nb);
\draw (and3.input 1) -- ++ (down: 0.25) -- (2.62, 0.8) (na);

\node (and4) [and gate, inputs=nn, rotate=90, scale=0.9] at (2.52,2.6) {};
\node (and5) [and gate, inputs=nn, rotate=90, scale=0.9] at (3.6,2.6) {};

\draw (or2.input 1) -- (and4) node[pos=0.6, left, color=sanae5] {$0.5$};
\draw (or2.input 2) -- ++ (down: 0.25) -| (and5) node[pos=0.4, above right, color=sanae5] {$0.3$};

\node (or4) [or gate, inputs=nn, rotate=90, scale=0.9] at (2.52,1.4) {};
\node (or5) [or gate, inputs=nn, rotate=90, scale=0.9] at (4.9,1.4) {};

\draw (or4.input 1) -- (a);
\draw (or4.input 2) -- ++ (down: 0.50) -| (na);

\draw (or5.input 1) -- ++ (down: 0.25) -| (b);
\draw (or5.input 2) -- ++ (down: 0.25) -- (5.9, 0.9) (nb);

\draw (and4.input 1) -- (2.43, 1.8) (or4) node[pos=0.4, left, color=sanae5] {$1$};
\draw (and4.input 2) -- ++ (down: 0.40) --++ (right: 1.40) --++ (down:0.9) --++ (right: 0.79)(b);

\draw (and5.input 1) -- (3.51, 1.0) -- (2.43, 1.0) ;
\draw (and5.input 2) -- ++ (down: 0.25) -| (or5) node[pos=0.4, above right, color=sanae5] {$1$};
\end{tikzpicture}
}
\end{subfigure}
\begin{subfigure}[b]{0.20\textwidth}
\centering
\includegraphics[scale=0.30, clip]{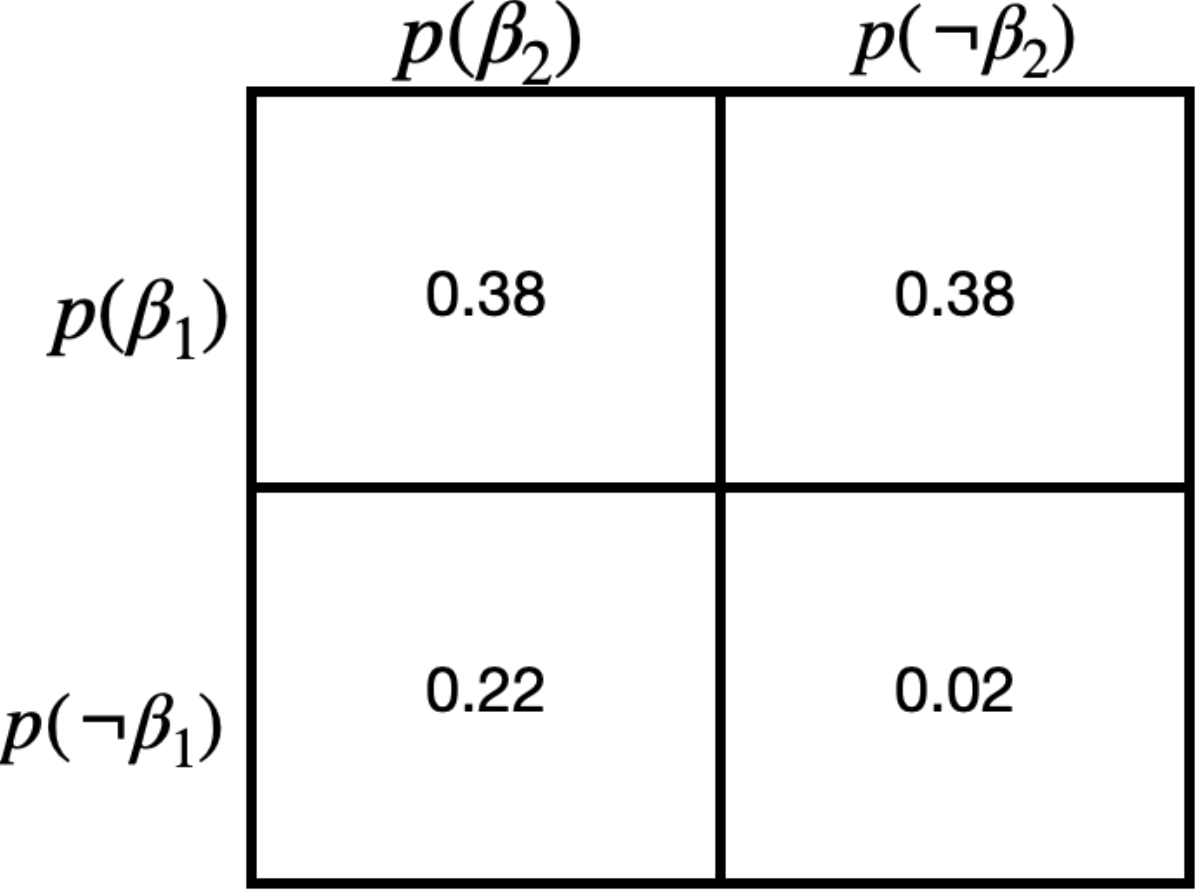}
\vspace{2em}
\end{subfigure}
\caption{%
(Left) Example of two compatible constraint circuits parameterized by the outputs of a neural network.
To compute the probability of a circuit, we plug in the output of the neural network $\vp_i$ and $1-\vp_i$ for positive and negative literal $i$, respectively.
The computation proceeds bottom-up, taking products at AND gates and summations at OR gates, and the probability is accumulated at the root of the circuit.
(Right) the conjunction of the two constraint circuits, its probability, computing the probabilities required for the mutual information using the law of total probability.
}
\label{fig:circuits_mi}
\end{figure*}

However, as discussed in \cref{sec:problem_statement}, training the neural network to satisfy
the local constraints can often be problematic: two \emph{dependent} constraints assumed
independent can often disagree on the value of their shared variables leading to opposing gradients. 
If we are afforded more computational resources, we can start  strengthening our approximation
by relaxing some of the independence assumptions made in our model.

\subsection{Deriving the Criterion}
The question then becomes, \emph{which independence assumptions to relax}.
We are, of course, interested in relaxing the independence assumptions that 
have the most positive impact on the quality of the approximation.
Or, put differently, we are interested in relaxing the independence assumptions 
for which we incur the most penalty for assuming, otherwise dependent constraints,
to be independent.
For each pair of constraints $\beta_i$ and $\beta_j$, for all $i \neq j$,
this corresponds to the Kullback-Leibler divergence of the product of their marginals
from their joint distribution, and is a measure of the modeling error we incur, in bits, 
by assuming the independence of the two constraints
\begin{equation}\label{eqn:kl_div}
    \KL\!\left(\prob_{(X,Y)} \Vert \prob_X \cdot \prob_Y\right)
\end{equation}
where $X$ and $Y$ are Bernoulli random variables, $X \sim \prob(\beta_i)$, $Y \sim \prob
(\beta_j)$, and $(X,Y) \sim \prob(\beta_i, \beta_j)$, 
for 
all $i,j$ such that $i \neq j$. 
\cref{eqn:kl_div} equivalently corresponds to the \emph{mutual information} $I(X;Y)$
given by
\begin{equation}\label{eqn:mi}
    I(X;Y) = \E_{(X,Y)}\!\left[ \log \frac{\prob_{(X,Y)}(X,Y)}{\prob_X(X)\cdot\prob_Y(Y)}\right],
\end{equation}
between the random variables $X$ and $Y$, or the measure of \emph{dependence} between 
them.
Intuitively, mutual information captures the information shared between $X$ and $Y$:
it measures how much knowing one reduces about the uncertainty of the other. When 
they are independent, then knowing one does not give any information about the
other, and therefore the mutual information is $0$. At the other extreme, one is a
deterministic function of the other, and therefore, the mutual information is 
maximized and equals to their entropy.
Note that the expectations in both \cref{eqn:kl_div} and \cref{eqn:mi} are over
the joint distribution $P_{(X,Y)}$.

We would be remiss, however, to dismiss the features learned by the
network, as they already encode some of the dependencies between the
constraints, affording us the ability to make stronger approximations.
That is, we are interested in the mutual information between all pairs
of constraints $\beta_i, \beta_j$ \emph{conditioned} on the neural
network's features.
Let $\mathcal{D}$ be our data distribution, and $Z$ be a random 
variable distributed according to $D$,  we are interested in computing
\begin{align}\label{eqn:conditional_mi}
    &I(X;Y\mid Z) = \E_{Z}\!\left[\E_{(X,Y)\mid Z}\!\left[ \log \frac{\prob(x,y \mid \zs)}{\prob(x \mid \vz)\cdot\prob(y \mid \zs)}\right]\right]\\
    &=\E_{Z}\!\left[\sum_{x = 0}^{1}\sum_{y = 0}^{1} \prob(x, y \mid \zs)\!\left[ \log \frac{\prob(x,y \mid \zs)}{\prob(x \mid \vz)\cdot\prob(y \mid \zs)}\right]\right],
\end{align}
where, as is common place, we estimate the outer expectation using 
Monte Carlo sampling from the data distribution.

Perhaps rather surprisingly, not withstanding the expectation w.r.t
the data distribution, the quantity in \cref{eqn:conditional_mi} is
hard to compute.
This is not only due to the intractability of the probability, which
as we have already stated is \#P-hard in general, but also due to the
hardness of conjunction, in general.
Loosely speaking, one could have constraints $\beta_i$ and $\beta_j$
for which the probability computation, $\prob(\beta_i)$ and $\prob(\beta_j)$
is tractable, yet computing $\prob(\alpha)$, where once again $\alpha = 
\beta_i \land \beta_j$, is hard \citep{Yujia2016, Khosravi19}.
Intuitively, the hardness of conjunction comes from finding the intersection
of the satisfying assignments without enumeration.
We formalize this in \cref{sec:tractable_computation}.

\begin{figure*}[t]
\scalebox{0.95}
{
\begin{minipage}[t]{\dimexpr.5\textwidth-.5\columnsep}
\begin{algorithm}[H]
\begin{algorithmic}
    \State \hskip-1em\textbf{Input}: Two compatible constraint circuits $\beta_1$ and $\beta_2$
    \State \hskip-1em\textbf{Output}: Mutual Information of $\beta_1$ and $\beta_2$ given features
    \\\textcolor{ourspecialtextcolor}{// Conjoin $\beta_1$ and $\beta_2$}
    \State $\alpha = \beta_1 \land \beta_2$
    \\\textcolor{ourspecialtextcolor}{// Compute the probability of $\alpha, \beta_1$ and $\beta_2$ c.f.\ \cref{fig:circuits_mi}}
    \State $p_\alpha, p_{\beta_1}, p_{\beta_2}$ = $\text{prob}(\alpha)$, $\text{prob}(\beta_1)$, $\text{prob}(\beta_2)$

    \\\textcolor{ourspecialtextcolor}{// Calculate marginals and joint using total probability}
    \State $p_{\X} = [1 - p_{\beta_1}, p_{\beta_1}]$, $p_{\Y} = [1 - p_{\beta_2}, p_{\beta_2}]$
    \State $p_{(\X, \Y)} = [[1 - p_{\beta_1} - p_{\beta_2} - p_{\alpha} ,p_{\beta_2} - p_\alpha], [p_{\beta_1} - p_\alpha, p_\alpha]]$
    
    \State $\text{mi} = 0$
    \For{$x$, $y$ \textbf{in} $\text{product}([0,1])$}
        \State $\text{mi} \mathrel{+}= p_{(\X, \Y)}[x][y] \times \log(\frac{p_{(\X, \Y)}[x][y]}{p_{\X}[x] \times p_{\Y}[y]})$
    \EndFor
    \vspace{-0.8mm}
    \State \textbf{return} $\text{mi}$
\end{algorithmic}
\caption{$\MI(\beta_1;\beta_2 \mid f(\xs))$}
\label{algo:MI}
\end{algorithm}
\end{minipage}}\hfill
\scalebox{0.95}
{
\begin{minipage}[t]{\dimexpr.5\textwidth-.5\columnsep}
\begin{algorithm}[H]
\begin{algorithmic}
    \State \hskip-1em\textbf{Input}: Current set of constraint circuits
    \State  \hskip-1em\textbf{Output}: \emph{Strengthened} set of constraints
    \State $\text{pwmi} = [\ ]$
    \For{$\beta_1$, $\beta_2$ \textbf{in} $\text{product}(\text{constraints})$}
        \State \textbf{if} $\text{disjoint}(\vars(\beta_i), \vars(\beta_j))$ \textbf{then} \textbf{continue}
        \\\textcolor{ourspecialtextcolor}{// Keep track of constraints with mutual information}
        \State $\text{pwmi}.\text{append}((\MI(\beta_i, \beta_j), \beta_i, \beta_j))$
    \EndFor
    \\\textcolor{ourspecialtextcolor}{// Consider only the top $\kappa$ pairs of constraints}
    \State $\text{to\_merge} = \text{sorted}(\text{pwmi, reverse=True})[:\kappa]$
    
    \For{$\text{mi}, \beta_1$, $\beta_2$ \textbf{in} $(\text{to\_merge})$}
        \State constraints.remove($\beta_i$, $\beta_j$)
        \State constraints.append($\beta_i \land \beta_j$)
    \EndFor
    
    \State \textbf{return} constraints
\end{algorithmic}
\caption{$\SemStr(\text{constraints}, \kappa)$}
\label{algo:Strengthen}
\end{algorithm}
\end{minipage}
}
\end{figure*}

\subsection{The Semantic Strengthening Algorithm}
For the purposes of this section, we will assume we can tractably
compute the conditional mutual information in \cref{eqn:conditional_mi},
and proceed with giving our Semantic Strengthening algorithm.
The idea is, simply put, to use the neural network to guide the process
of relaxing the independence assumptions introduced between the constraints.
Specifically, we are given an interval, $\eta$, a constraint budget, $\kappa$,
and a computational budget $\tau$.
We initiate the process of training the neural network, interrupting training
every $\eta$ epochs, computing the conditional mutual information between
pairs of constraints,
considering only those pairs sharing at least one variable (e.g.
the two constraints asserting the uniqueness of the first and last row, respectively,
do not share variables, are therefore independent, and by definition have a mutual 
information of $0$, so we need not consider joining them, \emph{yet}).
Subsequently, we identify the $\kappa$ pairs of constraints with the highest pairwise
conditional mutual information, and that therefore, have the most detrimental
effect on the quality of our approximation.
We detect the strongly connected components of constraints, and conjoin them: 
if $\beta_1$ and $\beta_2$ should be made dependent, and $\beta_2$ and $\beta_3$ 
should be made dependent, then $\beta_1$ , $\beta_2$ and $\beta_3$ are made dependent.
We delete the old constraints from, and add the new constraints, to our set of constraints,
and resume training.
This process is repeated every $\eta$ epochs until we have exhausted our computational budget $\tau$.
Our full algorithm is shown in \cref{algo:Strengthen}.%

\subsection{Tractably Computing the Criterion}\label{sec:tractable_computation}
Unlike previous approaches~\citep{Chen2018, Mesner2019ConditionalMI, Tezuka2021InformationBA}, we do not need to resort to variational approximations or neural estimation to
compute the mutual information, and instead appeal to the language of \emph{tractable
circuits}.
That is, we appeal to knowledge compilation techniques\textemdash a class of methods
that transform, or \emph{compile}, a logical theory into a target form, \emph{tractable
circuits}, which represent functions as parameterized computational graphs.
By imposing certain structural properties on these computational graphs, we enable
the tractable computation of certain classes of probabilistic queries over the
encoded functions.
As such, circuits provide us with a language for building and reasoning about tractable 
representations.

\paragraph{Logical Circuits} 
More formally, a \emph{logical circuit} is a directed, acyclic computational 
graph representing a logical formula.
Each node $n$ in the DAG encodes a logical sub-formula, denoted $[n]$.
Each inner node in the graph is either an AND or an OR gate, and each leaf 
node encodes a Boolean literal ($Y$ or $\lnot Y$). 
We denote by $\ch(n)$ the set of $n$'s children, that is, the operands of 
its logical gate.

\paragraph{Structural Properties}  As already alluded to, circuits enable 
the tractable computation of certain classes of queries over encoded functions
granted that a set of structural properties are enforced. We explicate such 
properties below.

A circuit is \emph{decomposable} if the inputs of every AND gate depend on 
disjoint sets of variables i.e.\ for $\alpha = \beta \land \gamma$, $\vars
(\beta) \cap \vars(\gamma) = \varnothing$.
Intuitively, decomposable AND nodes encode local factorizations over
variables of the function. For simplicity, we assume
that decomposable AND gates always have two inputs, a condition that
can be enforced on any circuit in exchange for a polynomial increase
in its size~\citep{vergari2015simplifying,peharz2020einsum}.

A second useful property is \emph{smoothness}.
A circuit is \emph{smooth} if the children
of every OR gate depend on the same set of variables i.e. for 
$\alpha = \bigvee_i \beta_i$, we have that $\vars(\beta_i) = 
\vars(\beta_j)\ \forall i,j$. Decomposability and smoothness 
are a sufficient and necessary condition for tractable integration
over arbitrary sets of variables in a single pass, as they allow 
larger integrals to decompose into smaller ones~\citep{choi2020pc}.

Furthermore, a circuit is said to be  \emph{deterministic} if, 
for any input, at most one child of every OR node has a non-zero
output i.e. for $\alpha = \bigvee_i \beta_i$, we have that $ \beta_i 
\land \beta_j = \bot$ for all $i \neq j$.
Similar to decomposability, determinism induces a recursive 
partitioning of the function, but over the support, i.e.\ 
satisfying assignments, of the function, rather than the
variables.
Determinism, taken together with smoothness and decomposability, 
allows us to tractably compute the probability of a constraint
~\citep{darwiche02}.%

What remains, is to show that we can tractably conjoin two constraints.
Conjoining two decomposable and deterministic circuits is NP-hard if we
wish the result to also be decomposable and deterministic, which as we
mentioned is a requirement for tractable probability 
computation~\citep{darwiche02, Yujia2016, Khosravi19}.
To guarantee the tractability of the probability computation of the 
conjoined constraint, we will, therefore, need to introduce one last
structural property, namely the notion of \emph{compatibility} between
two circuits~\citep{VergariNeurIPS21}.
Two circuits, $c_1$ and $c_2$ over variables $\Ys$ are said to be compatible
if (1) they are smooth and decomposable, and (2) any pair of AND nodes, $n 
\in c_1$ and $m \in c_2$ with the same scope over $\Ys$ can be rearranged to be
mutually compatible and decompose in the same way i.e.\ $\vars(n) = \vars(m) 
\implies \vars(n_i) = \vars(m_i)$, and $n_i$ and $m_i$ are compatible, for some
arrangement of the inputs $n_i$ and $m_i$ of $n$ and $m$.
A sufficient condition for compatibility is that both $c_1$ and $c_2$ share the
exact same hierarchical scope partitioning~\citep{VergariNeurIPS21},
sometimes called a vtree or variable ordering~\citep{choi2020pc,pipatsrisawat2008new}.
Intuitively, the two circuits should share the order in which they factorize the 
function over its variables.
\cref{fig:circuits_mi} shows an example of smooth, decomposable, deterministic and compatible circuits.

At a high level, there exist off-the-shelf compilers utilizing SAT solvers, 
essentially through case analysis, to compile a logical formula into a tractable
logical circuit.
We are agnostic to the exact flavor of circuit so long as the properties outlined
herein are respected.
In our experiments, we use PySDD\footnote{https://github.com/wannesm/PySDD} --
a Python SDD compiler~\citep{darwiche11,choi2013dynamic}.

Now that we have shown that we can tractably compute the probabilities
$\prob(\beta_1), \prob(\beta_2)$ and $\prob(\alpha)$, we can utilize
the law of total probability (c.f.\ \cref{fig:circuits_mi}) to compute
the remaining probabilities, and therefore, the mutual information.
Our algorithm is shown in \cref{algo:MI}.

\section{Related Work}\label{sec:related_work}

There has been increasing interest in combining neural learning with symbolic reasoning,
a class of methods that has been termed \emph{neuro-symbolic} methods, studying how to best combine both paradigms in a bid to accentuate their positives and mitigate their negatives.
The focus of many such approaches has therefore been on making probabilistic reasoning tractable through first-order approximations, and differentiable, through reducing logical formulas into arithmetic objectives, replacing logical operators with their fuzzy t-norms, and implications with inequalities~\citep{kimmig2012short,rocktaschel2015,fischer19a, pryor22}.

\citet{diligenti2017} and \citet{donadello2017} use first-order logic to specify constraints on outputs of a neural network. They employ fuzzy logic to reduce logical formulas into differential, arithmetic objectives denoting the extent to which neural network outputs violate the constraints, thereby supporting end-to-end learning under constraints. More recently, \citet{Xu18} introduced semantic loss, which circumvents the shortcomings of fuzzy approaches, while supporting end-to-end learning under constraints. More precisely, \emph{fuzzy reasoning} is replaced with \emph{exact probabilistic reasoning}, by compiling logical formulae into structures supporting efficient probabilistic queries.
\cite{LiuAAAI23} use semantic loss to simultaneously
learn a neural network and extract generalized logic rules. Different from other neural-symbolic methods that require background knowledge and candidate logical rules, they aim to induce task semantics with minimal priors.

Another class of neuro-symbolic approaches have their roots in logic programming. DeepProbLog~\citep{manhaeve2018} extends ProbLog, a probabilistic logic programming language, with the capacity to process neural predicates, whereby the network's outputs are construed as the probabilities of the corresponding predicates. This simple idea retains all essential components of ProbLog: the semantics, inference mechanism, and the implementation. \cite{KR2021-45} attempts to scale DeepProbLog by considering only the top-$k$ proof paths. In a similar vein, \citet{dai2018} combine domain knowledge specified as purely logical Prolog rules with the output of neural networks, dealing with the network's uncertainty through revising the hypothesis by iteratively replacing the output of the neural network with anonymous variables until a consistent hypothesis can be formed. \citet{bosnjak2017programming} present a framework combining prior procedural knowledge, as a Forth program, with neural functions learned through data. The resulting neural programs are consistent with specified prior knowledge and optimized with respect to data.

\begin{figure*}[th!]
\centering
\includegraphics[scale=0.175]{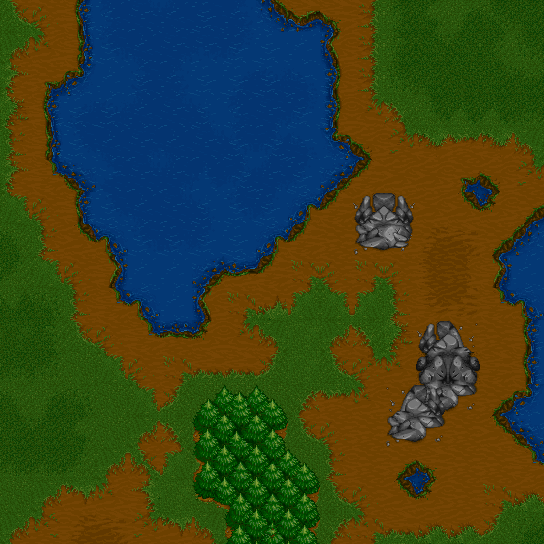}
\hfill
\includegraphics[scale=0.175]{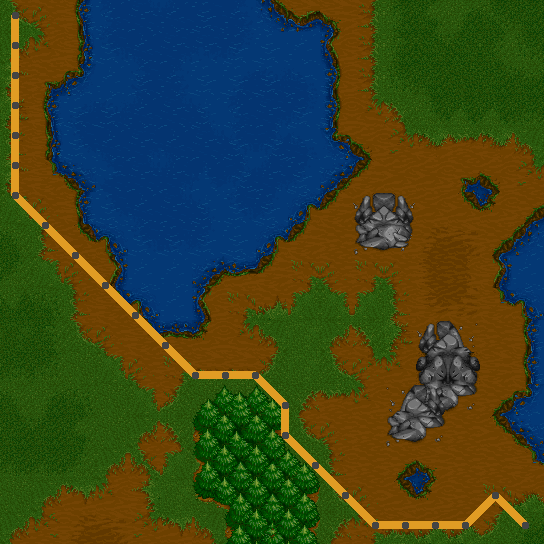}
\hfill
\includegraphics[scale=0.8]{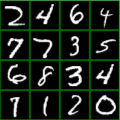}
\hfill
\includegraphics[scale=0.8]{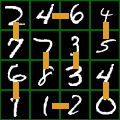}
\caption{An example of a Warcraft terrain map (left) and an MNIST grid, and the corresponding groundtruth labels.
}
\label{fig:example_tasks}
\end{figure*}

There has recently been a plethora of approaches ensuring consistency by embedding the constraints as predictive layers, including semantic probabilistic layers (SPLs) \citep{Ahmed2022}, MultiplexNet~\citep{Hoernle2021MultiplexNetTF} and HMCCN~\citep{giunchiglia2020coherent}.
Much like semantic loss \citep{Xu18}, SPLs maintain sound probabilistic semantics, and while displaying impressive scalability to real world problems, but might struggle with encoding harder constraints.
SIMPLE \citep{AhmedICLR2023} proposes an SPL for the $k$-subset distribution, to be used as a latent space to induce a distribution over features, for which they derive a low-bias, low-variance gradient estimator.
MultiplexNet is able to encode only constraints in disjunctive normal form, which is problematic for generality and efficiency as neuro-symbolic tasks often involve an intractably large number of clauses.
HMCCN encodes label dependencies as fuzzy relaxation and is the current state-of-the-art model for hierarchical mutli-label classification~\citep{giunchiglia2020coherent}, but, similar to its recent extension~\citep{giunchiglia2021multi}, is restricted to a certain family of constraints.
\cite{Daniele2022RefiningNN} discusses how to enforce the consistency for fuzzy relaxations with general formulas.

\section{Experimental Evaluation}\label{sec:experiments}
We evaluated our approach, semantic strengthening, on several neuro-symbolic tasks,
namely Warcraft minimum-cost path finding, minimum-cost perfect matching of MNIST 
digits, as well as the task of training neural networks to solve Sudoku puzzles.
The challenge with all of the above tasks, when looked at through a neuro-symbolic
lens, is the vastness of the state space: as previously mentioned, there are $6.6 
\times 10^{21}$ valid $9 \times 9$ Sudokus, and the number of valid matchings, 
or paths in a grid grows doubly-exponentially in the grid size\textemdash simply
too much to enumerate.
Even approaches like semantic loss which rely on circuit approaches to exploit the
local structure in the problem, essentially through caching solutions to repeated
subproblems, do not scale to large instances of these tasks.

As has been established in previous work~\citep{Xu18,Ahmed22nesyentropy,Ahmed2022},
label-level accuracy, or the accuracy of predicting individual labels is very often
a poor indication of the performance of the neural network, and is often uninteresting
in neuro-symbolic settings, where we are rather more interested in the accuracy of 
our predicted structure object \emph{exactly} matching the ground truth, e.g., \emph{is the prediction a shortest path?}, a metric which we denote ``Exact'' in our
experiments, as well as the accuracy of predicting objects that are \emph{consistent} 
with the constraint, e.g., \emph{is the prediction a valid path?}, a metric which we 
denote ``Consistent'' in our experiments.
Note that, unlike the other two tasks, for the case of Sudoku, 
these measures are one and the same: a valid Sudoku has a single \emph{unique} solution.

In all of our experiments, we compare against two baselines: a neural network, whose architecture
we specify in the corresponding experimental section, and the same neural network augmented
with product t-norm, where we assume the independence of  constraints throughout training.

\paragraph{Warcraft Shortest Path} 
We evaluate our approach, semantic strengthening, on the challenging task of predicting
the minimum-cost path in a weighted grid imposed over Warcraft terrain maps.
Following \citet{Pogancic2020}, our training set consists of $10,000$
terrain maps curated using the Warcraft II tileset.
Each map encodes an underlying grid of dimension $12 \times 12$, where each vertex is 
assigned a cost depending on the type of terrain it represents (e.g. earth has lower cost than water).
The shortest (minimum cost) path between the top left and bottom right vertices is encoded as
an indicator matrix, and serves as label. 
\cref{fig:example_tasks} shows an example input presented to the network and the input with an annotated shortest path as a groundtruth.
Presented with an image of a terrain map, a convolutional neural network\textemdash similar to \citet{Pogancic2020}, we use ResNet18 \citep{He2016}\textemdash outputs a $12 \times 12$ binary matrix indicating a set of vertices.
Note that the minimum-cost path is not unique: there may exist several paths sharing the same minimum cost, all of which are considered to be
correct by our metrics.
Table~\ref{tab:sp} shows our results.

\begin{table}[h]
\centering
\caption {Warcraft shortest path prediction results}
{
\begin{tabular}{@{}l l l l @{}}
Test accuracy \%  & & Exact & Consistent\\
\midrule \midrule
ResNet-18&   & $44.80$  & $56.90$\\
\midrule
+ Product t-norm&  &  $50.40$ & $63.20$\\
+ Semantic Strengthening&  &  $\mathbf{61.20}$&  $\mathbf{72.70}$\\
\end{tabular}
}
\label{tab:sp}
\end{table}

We observe that incorporating constraints into learning improves the accuracy of predicting
the optimal path from $44.80\%$ to $50.40\%$, and the accuracy of predicting a \emph{valid}
path from $56.90\%$ to $63.20\%$, as denoted by the ``Exact'' and ``Consistent'' metrics,
respectively. Furthermore, and perhaps more interestingly, we see that our approach, 
\emph{semantic strengthening}, greatly improves upon the baseline, as well as product t-norm
improving the accuracy of predicting the optimal path from $44.80\%$ and $50.40\%$ to $61.20\%$, 
while greatly improving the accuracy of predicting a valid path from $56.90\%$ and $63.20\%$ to 
$72.70\%$.

\paragraph{MNIST Perfect Matching}
Our next task consists in predicting a minimum-cost perfect-matching of a set
of $k^2$ MNIST digits arranged in a $k \times k$ grid, where diagonal matchings
are not permitted.
We consider the problem for the instance when $k = 10$.
Similar to \citet{Pogancic2020}, we generate the ground truth
by considering the underlying $k \times k$ grid graph, and solving
a min-cost perfect-matching problem using Blossom V \citep{Kolmogorov2009},
where the edge weights are given simply by reading the two vertex digits as
a two-digit number, reading downwards for vertical edges, and left to right
for horizontal edges.
The minimum-cost perfect matching label is then encoded
as an indicator vector for the subset of the selected edges.
Similar to the Warcraft experiment, the grid image is input to a (pretrained)
ResNet-18, which simply outputs a set of predicted edges.
\cref{tab:pm} shows our results.

\begin{table}[h]
\centering
\caption {Perfect Matching prediction test results}
{
\begin{tabular}{@{}l l l l @{}}
Test accuracy \%  & & Exact & Consistent\\
\midrule \midrule
ResNet-18&   & $9.30$  & $10.00$\\
\midrule
+ Product t-norm&  &  $12.70$ & $12.90$\\
+ Semantic Strengthening&  &  $\mathbf{15.50}$&  $\mathbf{18.40}$\\
\end{tabular}
}
\label{tab:pm}
\end{table}

Similar to the Warcraft experiment, we observe that incorporating constraints into learning 
improves the accuracy of predicting the optimal perfect matching from $9.30\%$ to $12.70\%$, and the 
accuracy of predicting a \emph{valid} perfect matching from $10.00\%$ to $12.90\%$, as denoted by the 
``Exact'' and ``Consistent'' metrics, respectively. Furthermore,
we see that our approach, \emph{semantic strengthening}, greatly improves upon the baseline, 
as well as product t-norm improving the accuracy of predicting the optimal perfect matching from $09.30\%$
and $12.70\%$ to $15.50\%$, while greatly improving the accuracy of predicting a valid perfect matching 
from $10.00\%$ and $12.90\%$ to $18.40\%$.

\paragraph{Sudoku}
Lastly, we consider the task of predicting a solution to a given Sudoku puzzle.
Here the task is, given a $9 \times 9$ partially-filled grid of numbers
to fill in the remaining cells in the grid such that the entries each row, column, 
and $3 \times 3$ square are unique i.e.\ each of the numbers from $1$ through $9$
appears exactly once.

We use the dataset provided by \citet{Wang19}, consisting of 10K Sudoku puzzles,
split into 9K training examples, and 1K test samples, all puzzles having 10 missing entries.

As our baseline, we follow \citet{Wang19} in using a convolutional neural network
modeled on that of \citet{Park2018}.
The input to the neural network is given as a bit representation
of the initial Sudoku board, along with a mask representing the bits
to be learned, i.e.\ the bits in the empty Sudoku cells.
The network interprets the bit inputs as 9 input image channels (one for each square
in the board) and uses a sequence of 10 convolutional layers (each with 512 3$\times$3
filters) to output the solution, with the mask input as a set of additional image channels
in the same format as the board.
\cref{tab:sudoku} shows our results.

\begin{table}[h]
\centering
\caption {Sudoku test results}
{
\begin{tabular}{@{}l l l l @{}}
Test accuracy \%  & & Exact & Consistent\\
\midrule \midrule
10-Layer ConvNet&   & $16.80$  & $16.80$\\
\midrule
+ Product t-norm&  &  $22.10$ & $22.10$\\
+ Semantic Strengthening&  &  $\mathbf{28.00}$&  $\mathbf{28.00}$\\
\end{tabular}
}
\label{tab:sudoku}
\end{table}

In line with our previous experiments, we observe that incorporating 
constraints into learning improves the accuracy of predicting correct
Sudoku solutions, the ``Exact'' metric from $16.80\%$ to $22.10\%$.
Furthermore, we see that our approach, \emph{semantic strengthening},
greatly improves upon the baseline, as well as product t-norm, improving
the accuracy from $16.80\%$  and $22.10\%$ to $28.00\%$.

\section{Conclusion}\label{sec:conclusion}
In conclusion, we proposed semantic strengthening, a tractable approach to neuro-symbolic learning,
that remains faithful to the probabilistic semantic of the distribution defined by the neural network
on a given constraint.
Semantic strengthening starts by assuming the independence of the clauses in a given constraint, thereby
reducing the, often intractable, problem of satisfying a global constraint, to the tractable problem of
satisfying individual local constraints.
It uses a principled criterion, conditional mutual information, to determine, and relax any unjustified
independence assumptions most detrimental to the quality of our approximation.
We have shown that we are able to greatly improve upon the baselines on three challenging tasks, where
semantic strengthening was able to increase the \emph{accuracy} and \emph{consistency} of the model's predictions.

\subsubsection*{Acknowledgements}
KA would like to thank Arthur Choi and Yoojung Choi for helpful discussions throughout the project.
This work was funded in part by the DARPA Perceptually-enabled Task Guidance (PTG) Program under
contract number HR00112220005, and NSF grants \#IIS-1943641, \#IIS-1956441, and \#CCF-1837129.
\bibliography{references}

\begin{thebibliography}{}

\bibitem[Ahmed et~al., 2022a]{Ahmed22pylon}
Ahmed, K., Li, T., Ton, T., Guo, Q., Chang, K.-W., Kordjamshidi, P., Srikumar,
  V., Van~den Broeck, G., and Singh, S. (2022a).
\newblock Pylon: A pytorch framework for learning with constraints.
\newblock In {\em Proceedings of the 36th AAAI Conference on Artificial
  Intelligence (Demo Track)}.

\bibitem[Ahmed et~al., 2022b]{Ahmed2022}
Ahmed, K., Teso, S., Chang, K.-W., Van~den Broeck, G., and Vergari, A. (2022b).
\newblock Semantic probabilistic layers for neuro-symbolic learning.
\newblock In {\em NeurIPS}.

\bibitem[Ahmed et~al., 2022c]{Ahmed22nesyentropy}
Ahmed, K., Wang, E., Chang, K.-W., and Van~den Broeck, G. (2022c).
\newblock Neuro-symbolic entropy regularization.
\newblock In {\em The 38th Conference on Uncertainty in Artificial
  Intelligence}.

\bibitem[Ahmed et~al., 2023]{AhmedICLR2023}
Ahmed, K., Zeng, Z., Niepert, M., and {Van den Broeck}, G. (2023).
\newblock {SIMPLE}: A gradient estimator for k-subset sampling.
\newblock In {\em ICLR}.

\bibitem[Asai and Hajishirzi, 2020]{asai2020}
Asai, A. and Hajishirzi, H. (2020).
\newblock Logic-{{Guided Data Augmentation}} and {{Regularization}} for
  {{Consistent Question Answering}}.
\newblock In {\em ACL}.

\bibitem[Bach et~al., 2017]{Bach2017}
Bach, S.~H., Broecheler, M., Huang, B., and Getoor, L. (2017).
\newblock Hinge-loss markov random fields and probabilistic soft logic.
\newblock {\em JMLR}.

\bibitem[Bo{\v{s}}njak et~al., 2017]{bosnjak2017programming}
Bo{\v{s}}njak, M., Rockt{\"a}schel, T., Naradowsky, J., and Riedel, S. (2017).
\newblock Programming with a differentiable forth interpreter.
\newblock In {\em Proceedings of the 34th ICML}.

\bibitem[Chen et~al., 2018]{Chen2018}
Chen, J., Song, L., Wainwright, M., and Jordan, M. (2018).
\newblock Learning to explain: An information-theoretic perspective on model
  interpretation.
\newblock In Dy, J. and Krause, A., editors, {\em Proceedings of the 35th
  International Conference on Machine Learning}, volume~80 of {\em Proceedings
  of Machine Learning Research}, pages 883--892. PMLR.

\bibitem[Choi and Darwiche, 2013]{choi2013dynamic}
Choi, A. and Darwiche, A. (2013).
\newblock Dynamic minimization of sentential decision diagrams.
\newblock In {\em Proceedings of the Twenty-Seventh AAAI Conference on
  Artificial Intelligence}, AAAI'13, page 187–194. AAAI Press.

\bibitem[Choi et~al., 2020]{choi2020pc}
Choi, Y., Vergari, A., and Van~den Broeck, G. (2020).
\newblock Probabilistic circuits: A unifying framework for tractable
  probabilistic modeling.

\bibitem[Dai et~al., 2018]{dai2018}
Dai, W.-Z., Xu, Q.-L., Yu, Y., and Zhou, Z.-H. (2018).
\newblock Tunneling neural perception and logic reasoning through abductive
  learning.

\bibitem[Daniele et~al., 2022]{Daniele2022RefiningNN}
Daniele, A., van Krieken, E., Serafini, L., and Harmelen, F.~V. (2022).
\newblock Refining neural network predictions using background knowledge.
\newblock {\em ArXiv}, abs/2206.04976.

\bibitem[Darwiche, 2011]{darwiche11}
Darwiche, A. (2011).
\newblock Sdd: A new canonical representation of propositional knowledge bases.
\newblock In {\em IJCAI}.

\bibitem[Darwiche and Marquis, 2002]{darwiche02}
Darwiche, A. and Marquis, P. (2002).
\newblock A knowledge compilation map.
\newblock {\em JAIR}.

\bibitem[De~Raedt et~al., 2020]{RaedtIJCAI2020}
De~Raedt, L., Dumančić, S., Manhaeve, R., and Marra, G. (2020).
\newblock From statistical relational to neuro-symbolic artificial
  intelligence.
\newblock In {\em IJCAI}.

\bibitem[Diligenti et~al., 2017]{diligenti2017}
Diligenti, M., Gori, M., and Saccà, C. (2017).
\newblock Semantic-based regularization for learning and inference.
\newblock {\em Artificial Intelligence}.

\bibitem[Donadello et~al., 2017]{donadello2017}
Donadello, I., Serafini, L., and d'Avila Garcez, A. (2017).
\newblock Logic tensor networks for semantic image interpretation.
\newblock In {\em IJCAI}.

\bibitem[Felgenhauer and Jarvis, 2005]{Felgenhauer2005}
Felgenhauer, B. and Jarvis, F. (2005).
\newblock Enumerating possible sudoku grids.

\bibitem[Fischer et~al., 2019]{fischer19a}
Fischer, M., Balunovic, M., Drachsler-Cohen, D., Gehr, T., Zhang, C., and
  Vechev, M. (2019).
\newblock {DL}2: Training and querying neural networks with logic.
\newblock In {\em ICML}.

\bibitem[Giunchiglia and Lukasiewicz, 2020]{giunchiglia2020coherent}
Giunchiglia, E. and Lukasiewicz, T. (2020).
\newblock Coherent hierarchical multi-label classification networks.
\newblock {\em Advances in Neural Information Processing Systems},
  33:9662--9673.

\bibitem[Giunchiglia and Lukasiewicz, 2021]{giunchiglia2021multi}
Giunchiglia, E. and Lukasiewicz, T. (2021).
\newblock Multi-label classification neural networks with hard logical
  constraints.
\newblock {\em Journal of Artificial Intelligence Research}, 72:759--818.

\bibitem[He et~al., 2016]{He2016}
He, K., Zhang, X., Ren, S., and Sun, J. (2016).
\newblock Deep residual learning for image recognition.
\newblock In {\em CVPR}.

\bibitem[Hoernle et~al., 2022]{Hoernle2021MultiplexNetTF}
Hoernle, N., Karampatsis, R.-M., Belle, V., and Gal, Y. (2022).
\newblock Multiplexnet: Towards fully satisfied logical constraints in neural
  networks.
\newblock In {\em AAAI}.

\bibitem[Khosravi et~al., 2019]{Khosravi19}
Khosravi, P., Choi, Y., Liang, Y., Vergari, A., and Van~den Broeck, G. (2019).
\newblock On tractable computation of expected predictions.
\newblock In {\em Advances in Neural Information Processing Systems 32
  (NeurIPS)}.

\bibitem[Kimmig et~al., 2012]{kimmig2012short}
Kimmig, A., Bach, S., Broecheler, M., Huang, B., and Getoor, L. (2012).
\newblock A short introduction to probabilistic soft logic.
\newblock In {\em Proceedings of the NIPS Workshop on Probabilistic
  Programming: Foundations and Applications}.

\bibitem[Kolmogorov, 2009]{Kolmogorov2009}
Kolmogorov, V. (2009).
\newblock Blossom v: a new implementation of a minimum cost perfect matching
  algorithm.
\newblock {\em Mathematical Programming Computation}, 1:43--67.

\bibitem[Li and Srikumar, 2019]{li2019}
Li, T. and Srikumar, V. (2019).
\newblock Augmenting neural networks with first-order logic.
\newblock In {\em ACL}.

\bibitem[Liu et~al., 2023]{LiuAAAI23}
Liu, A., Xu, H., Van~den Broeck, G., and Liang, Y. (2023).
\newblock Out-of-distribution generalization by neural-symbolic joint training.
\newblock In {\em Proceedings of the 37th AAAI Conference on Artificial
  Intelligence}.

\bibitem[Manhaeve et~al., 2018]{manhaeve2018}
Manhaeve, R., Dumancic, S., Kimmig, A., Demeester, T., and De~Raedt, L. (2018).
\newblock Deepproblog: Neural probabilistic logic programming.
\newblock In {\em NeurIPS}.

\bibitem[Manhaeve et~al., 2021]{KR2021-45}
Manhaeve, R., Marra, G., and De~Raedt, L. (2021).
\newblock {Approximate Inference for Neural Probabilistic Logic Programming}.
\newblock In {\em {Proceedings of the 18th International Conference on
  Principles of Knowledge Representation and Reasoning}}, pages 475--486.

\bibitem[Medina~Grespan et~al., 2021]{Grespan21}
Medina~Grespan, M., Gupta, A., and Srikumar, V. (2021).
\newblock Evaluating relaxations of logic for neural networks: A comprehensive
  study.
\newblock In Zhou, Z.-H., editor, {\em Proceedings of the Thirtieth
  International Joint Conference on Artificial Intelligence, {IJCAI-21}}, pages
  2812--2818. International Joint Conferences on Artificial Intelligence
  Organization.
\newblock Main Track.

\bibitem[Mesner and Shalizi, 2019]{Mesner2019ConditionalMI}
Mesner, O.~C. and Shalizi, C.~R. (2019).
\newblock Conditional mutual information estimation for mixed discrete and
  continuous variables with nearest neighbors.
\newblock {\em arXiv: Statistics Theory}.

\bibitem[Minervini et~al., 2017]{minervini2017}
Minervini, P., Demeester, T., Rockt{\"{a}}schel, T., and Riedel, S. (2017).
\newblock Adversarial sets for regularising neural link predictors.
\newblock In {\em {UAI}}.

\bibitem[Mullenbach et~al., 2018]{mullenbach2018explainable}
Mullenbach, J., Wiegreffe, S., Duke, J., Sun, J., and Eisenstein, J. (2018).
\newblock Explainable prediction of medical codes from clinical text.
\newblock {\em arXiv preprint arXiv:1802.05695}.

\bibitem[Park, 2018]{Park2018}
Park, K. (2018).
\newblock Can convolutional neural networks crack sudoku puzzles?
\newblock \url{https://github.com/Kyubyong/sudoku}.

\bibitem[Peharz et~al., 2020]{peharz2020einsum}
Peharz, R., Lang, S., Vergari, A., Stelzner, K., Molina, A., Trapp, M., Van~den
  Broeck, G., Kersting, K., and Ghahramani, Z. (2020).
\newblock Einsum networks: Fast and scalable learning of tractable
  probabilistic circuits.
\newblock In {\em International Conference of Machine Learning}.

\bibitem[Pipatsrisawat and Darwiche, 2008]{pipatsrisawat2008new}
Pipatsrisawat, K. and Darwiche, A. (2008).
\newblock New compilation languages based on structured decomposability.
\newblock In {\em AAAI}, volume~8, pages 517--522.

\bibitem[Pogančić et~al., 2020]{Pogancic2020}
Pogančić, M.~V., Paulus, A., Musil, V., Martius, G., and Rolinek, M. (2020).
\newblock Differentiation of blackbox combinatorial solvers.
\newblock In {\em ICLR}.

\bibitem[Pryor et~al., 2022]{pryor22}
Pryor, C., Dickens, C., Augustine, E., Albalak, A., Wang, W.~Y., and Getoor, L.
  (2022).
\newblock Neupsl: Neural probabilistic soft logic.

\bibitem[Rockt{\"a}schel et~al., 2015]{rocktaschel2015}
Rockt{\"a}schel, T., Singh, S., and Riedel, S. (2015).
\newblock Injecting logical background knowledge into embeddings for relation
  extraction.
\newblock In {\em Proceedings of the 2015 Conference of the NAACL}.

\bibitem[Shen et~al., 2016]{Yujia2016}
Shen, Y., Choi, A., and Darwiche, A. (2016).
\newblock Tractable operations for arithmetic circuits of probabilistic models.
\newblock In Lee, D., Sugiyama, M., Luxburg, U., Guyon, I., and Garnett, R.,
  editors, {\em Advances in Neural Information Processing Systems}, volume~29.
  Curran Associates, Inc.

\bibitem[Strehl, 2001]{STREHL2001}
Strehl, V. (2001).
\newblock Counting domino tilings of rectangles via resultants.
\newblock {\em Advances in Applied Mathematics}, 27(2):597--626.

\bibitem[Tezuka and Namekawa, 2021]{Tezuka2021InformationBA}
Tezuka, T. and Namekawa, S. (2021).
\newblock Information bottleneck analysis by a conditional mutual information
  bound.
\newblock {\em Entropy}, 23.

\bibitem[Valiant, 1979]{Valiant1979b}
Valiant, L. (1979).
\newblock The complexity of computing the permanent.
\newblock {\em Theoretical Computer Science}.

\bibitem[van Krieken et~al., 2020]{Krieken2020AnalyzingDF}
van Krieken, E., Acar, E., and Harmelen, F.~V. (2020).
\newblock Analyzing differentiable fuzzy logic operators.
\newblock {\em ArXiv}, abs/2002.06100.

\bibitem[Vergari et~al., 2021]{VergariNeurIPS21}
Vergari, A., Choi, Y., Liu, A., Teso, S., and Van~den Broeck, G. (2021).
\newblock A compositional atlas of tractable circuit operations for
  probabilistic inference.
\newblock In {\em NeurIPS}.

\bibitem[Vergari et~al., 2015]{vergari2015simplifying}
Vergari, A., Di~Mauro, N., and Esposito, F. (2015).
\newblock Simplifying, regularizing and strengthening sum-product network
  structure learning.
\newblock In {\em Joint European Conference on Machine Learning and Knowledge
  Discovery in Databases}.

\bibitem[Wang et~al., 2019]{Wang19}
Wang, P., Donti, P.~L., Wilder, B., and Kolter, J.~Z. (2019).
\newblock Satnet: Bridging deep learning and logical reasoning using a
  differentiable satisfiability solver.
\newblock In {\em {ICML}}, volume~97 of {\em Proceedings of Machine Learning
  Research}, pages 6545--6554. {PMLR}.

\bibitem[Xu et~al., 2018]{Xu18}
Xu, J., Zhang, Z., Friedman, T., Liang, Y., and Van~den Broeck, G. (2018).
\newblock A semantic loss function for deep learning with symbolic knowledge.
\newblock In {\em Proceedings of the 35th {ICML} 2018}.

\end{thebibliography}
\end{document}